\documentclass{article}
\usepackage{geometry}
\usepackage{times}
\usepackage{xcolor}
\usepackage{natbib}
\usepackage{bm}
\usepackage{graphicx}
\usepackage{amssymb}
\usepackage{multirow}
\usepackage{hyperref}
\usepackage{booktabs}
\usepackage{amsmath}

\definecolor{darkblue}{rgb}{0, 0, 0.5}
\definecolor{darkred}{rgb}{1, 0, 0}

\hypersetup{
    colorlinks=true,
    linkcolor=darkred,
    anchorcolor=purple,
    filecolor=darkblue,
    urlcolor=darkblue,
    citecolor=darkblue
}

\def\argmax{\mathop{\rm argmax}}
\def\argmin{\mathop{\rm argmin}}

\newcommand*{\Bhline}[0]{\noalign{\global\setlength{\arrayrulewidth}{0.9pt}}
\hline\noalign{\global\setlength{\arrayrulewidth}{0.4pt}}}

\title{\textbf{Towards a Universal Continuous Knowledge Base}}
\author{Gang Chen$^{1,3,4}$, Maosong Sun$^{1,3,4, 5}$, and Yang Liu$^{1,2,3,4,5} \thanks{Yang Liu is the corresponding author:  \tt{liuyang2011@tsinghua.edu.cn}.}$
\\
\\
$^1$Department of Computer Science and Technology, Tsinghua University \\
$^2$Institute for AI Industry Research, Tsinghua University \\
$^3$Institute for Artificial Intelligence, Tsinghua University \\
$^4$Beijing National Research Center for Information Science and Technology \\
$^5$Beijing Academy of Artificial Intelligence}
\date{}

\begin{document}
\maketitle

\begin{abstract}
In artificial intelligence (AI), knowledge is the information required by an intelligent system to accomplish tasks. While traditional knowledge bases use discrete, symbolic representations, detecting knowledge encoded in the continuous representations learned from data has received increasing attention recently. In this work, we propose a method for building a continuous knowledge base (CKB) that can store knowledge imported from multiple, diverse neural networks. The key idea of our approach is to define an interface for each neural network and cast knowledge transferring as a function simulation problem. Experiments on text classification show promising results: the CKB imports knowledge from a single model and then exports the knowledge to a new model, achieving comparable performance with the original model. More interesting, we import the knowledge from multiple models to the knowledge base, from which the fused knowledge is exported back to a single model, achieving a higher accuracy than the original model. With the CKB, it is also easy to achieve knowledge distillation and transfer learning. Our work opens the door to building a universal continuous knowledge base to collect, store, and organize all continuous knowledge encoded in various neural networks trained for different AI tasks.
\end{abstract}

\section{Introduction}

A {\em knowledge base} (KB) is a centralized repository where structured and unstructured information is stored, organized, and shared. KBs have been widely used to accomplish artificial intelligence (AI) tasks. While their initial use was closely connected with expert systems \citep{Hayes-Roth:83}, knowledge graphs such as Freebase \citep{Bollacker:08} have become a typical example of knowledge base used in AI systems recently, showing their effectiveness in applications such as image classification \citep{Marino:17}, natural language processing \citep{Zhang:18}, and bioinformatics \citep{Ernst:15}.

While the use of \textit{discrete representations} enables KBs to be easily interpretable to humans, it is notoriously challenging to build such discrete knowledge bases. On one hand, it is time-consuming and labor-intensive to manually construct KBs \citep{Lenat:90}. The situation aggravates when it comes to a vertical domain for which only a limited number of annotators qualify to have the expertise. On the other hand, although machine learning approaches are able to mine knowledge from data automatically, they often inevitably have limitations in the correctness, depth, and breadth of knowledge they can acquire \citep{Richardson:03}. Therefore, how to build large-scale, high-quality, and wide-coverage knowledge bases still remains a grand challenge to the community.

The past two decades have witnessed the rapid progress of deep learning \citep{Hinton:06}, which has proven effective in learning {\em continuous representations} from data, leading to substantial improvements in a variety of AI tasks like speech recognition \citep{Dahl:11}, image classification \citep{Krizhevsky:12}, and machine translation \citep{Vaswani:17}. More recently, there has been a significant paradigm shift from learning continuous representations from limited labeled data in a supervised fashion to from abundant unlabeled data in a self-supervised way \citep{Devlin:19, Brown:20}, which further advances the development of learning continuous representations from data.

Given the remarkable success of deep learning, an interesting question naturally arises: {\em Is knowledge encoded in the learned continuous representations}? A number of researchers have developed probing methods to evaluate the extent to which continuous representations encode knowledge of interest \citep{Linzen:16, Belinkov:17, Blevins:18, Hewitt:19}. For example, while \citet{Belinkov:17} conduct a quantitive evaluation that sheds lights on the ability of neural machine translation models to capture word structure, \citet{Hewitt:19} propose a structural probe that can test whether syntax trees are consistently embedded in a linear transformation of word representation space.

\begin{figure}[!t]
    \centering
    \includegraphics[width=0.45\textwidth]{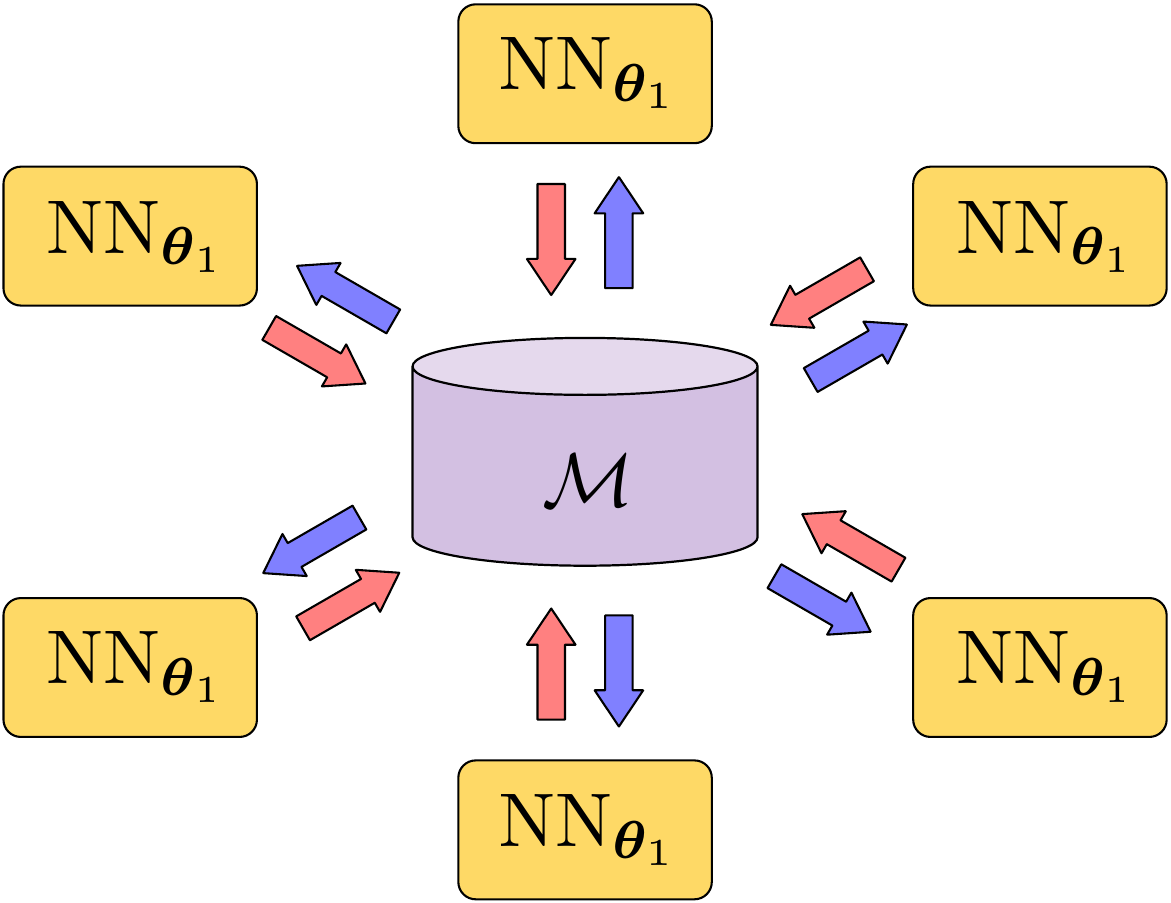}
    \caption{A universal continuous knowledge base. $\mathcal{M}$ denotes the knowledge base. $\mathrm{NN}_{\bm{\theta}_n}$ ($n \in [1, 6]$) denotes the $n$-th neural network parameterized by $\bm{\theta}_n$. On one hand, continuous knowledge encoded in a neural network can be imported to the knowledge base. On the other hand, continuous knowledge stored in the knowledge based can be exported to a neural network. Note that these neural networks are independent of each other: they can have different model structures and be trained for different AI tasks.
    } \label{fig:overview}
\end{figure}

If knowledge can be defined as the information required by a system to accomplish AI tasks, we would like to distinguish between two categories of knowledge: {\em discrete} and {\em continuous}. While discrete knowledge uses symbolic representations explicitly handcrafted by humans, continuous knowledge is implicitly encoded in neural networks automatically trained on data. If each trained neural network can be seen as a repository of continuous knowledge, there is an important question that needs to be answered: {\em Is it possible to build a universal continuous knowledge base that stores knowledge imported from diverse neural networks?}

In this work, we propose a method for building a universal continuous knowledge base (CKB). As shown in Figure~\ref{fig:overview}, the CKB allows for knowledge transferring between multiple, diverse neural networks. The knowledge encoded in one neural network can be imported to the CKB, from which the stored knowledge can be exported to another neural network. As a neural network can be seen as a parameterized function, we treat the function and its parameters as the continuous knowledge encoded in the neural network  (Section~\ref{sec:knowledge_encoded_in_a_neural_network}). Using a memory hierarchy to represent the knowledge base (Section~\ref{sec:design_of_a_continuous_knowledge_base}), our approach defines an interface function for each neural network and casts importing and exporting knowledge as a function simulation problem (Sections~\ref{sec:import_knowledge_from_a_neural_network} and~\ref{sec:export_knowledge_to_a_neural_network}). We adopt multi-task training to import knowledge from multiple neural networks (Section~\ref{sec:import_knowledge_from_multiple_neural_network}). Experiments on text classification show that our method is able to fuse knowledge imported from multiple, diverse neural networks and obtain better performance than single neural networks. It is also easy to use the knowledge base to simulate other learning paradigms such as knowledge distillation and transfer learning.

\section{Approach}

To use a continuous knowledge base to store the knowledge encoded in neural networks, we need to answer a number of fundamental questions:

\begin{enumerate}
\setlength{\itemsep}{4pt}
\setlength{\parskip}{0pt}
\setlength{\parsep}{0pt}
\item What is the knowledge encoded in a neural network? (Section~\ref{sec:knowledge_encoded_in_a_neural_network})
\item How to design a continuous knowledge base? (Section~\ref{sec:design_of_a_continuous_knowledge_base})
\item How to import the knowledge encoded in a neural network into the continuous knowledge base? (Section~\ref{sec:import_knowledge_from_a_neural_network})
\item How to import the knowledge encoded in multiple neural networks to the continuous knowledge base? (Section~\ref{sec:import_knowledge_from_multiple_neural_network})
\item How to export the knowledge stored in the continuous knowledge base to a neural network? (Section~\ref{sec:export_knowledge_to_a_neural_network})
\end{enumerate}

\subsection{Knowledge Encoded in a Neural Network} \label{sec:knowledge_encoded_in_a_neural_network}

\begin{figure}[!t]
\centering
\includegraphics[width=0.35\textwidth]{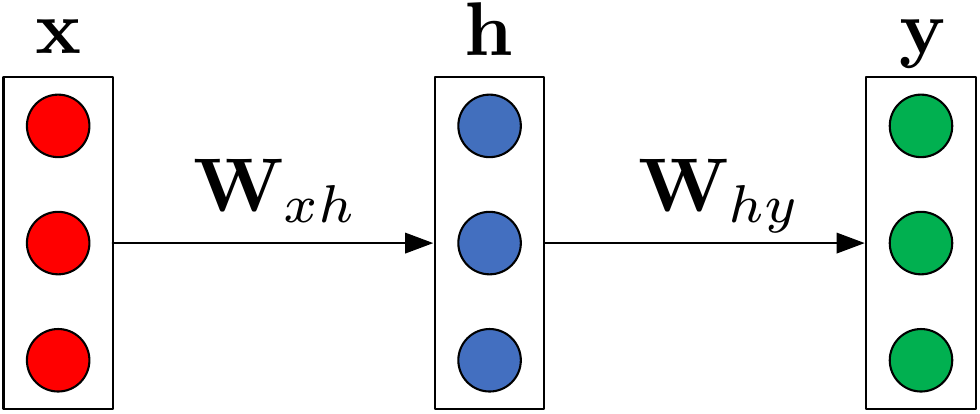}
\caption{Example of a feed-forward neural network that can be seen as a composite function. We treat the parameterized function as the continuous knowledge encoded in the neural network. $\mathbf{x}$ is the input, $\mathbf{h}$ is the hidden state, and $\mathbf{y}$ is the output. $\mathbf{W}_{xh}$ and $\mathbf{W}_{hy}$ are parameters.} \label{fig:fnn}
\end{figure}

A neural network can be seen as a parameterized, composite function. For example, Figure~\ref{fig:fnn} shows a simple feed-forward neural network involving the composition of two non-linear functions:
\begin{align}
\mathbf{h} &= f(\mathbf{W}_{xh}\mathbf{x}) \\
\mathbf{y} &= g(\mathbf{W}_{hy} \mathbf{h})
\end{align}
where $\mathbf{x}$ is the input layer, $\mathbf{h}$ is the hidden layer, $\mathbf{y}$ is the output layer, $\mathbf{W}_{xh}$ is the weight matrix between the input and hidden layers, $\mathbf{W}_{hy}$ is the weight matrix between the hidden and output layers, and $f(\cdot)$ and $g(\cdot)$ are two non-linear functions. For simplicity, we omit bias terms. The neural network shown in Figure~\ref{fig:fnn} can also be denoted by
\begin{eqnarray}
\mathbf{y} = \mathrm{FNN}_{\bm{\theta}}(\mathbf{x}) = g_{\bm{\theta}_2}( f_{\bm{\theta}_1}( \mathbf{x}))
\end{eqnarray}
where $\mathrm{FNN}_{\bm{\theta}}(\cdot)$ is a non-linear function parameterized by $\bm{\theta}=\{ \mathbf{W}_{xh}, \mathbf{W}_{hy} \}$, $f_{\bm{\theta}_1}(\cdot)$ is a non-linear function parameterized by $\bm{\theta}_1=\{\mathbf{W}_{xh}\}$, and $g_{\bm{\theta}_2}(\cdot)$ is a non-linear function parameterized by $\bm{\theta}_2 = \{ \mathbf{W}_{hy}\}$.

We distinguish between two functions related to neural networks:
\begin{enumerate}
\setlength{\itemsep}{4pt}
\setlength{\parskip}{0pt}
\setlength{\parsep}{0pt}
\item Global function: a function that maps the input of a neural network to its output.
\item Local function: a function that participates in the composition of a global function.
\end{enumerate}
For example, $\mathrm{FNN}_{\bm{\theta}}(\cdot)$ is a global function and $f_{\bm{\theta}_1}(\cdot)$ is a local function.

Given a trained neural network that is able to accomplish an AI task, we believe that it is the {\em parameterized function} that represents the continuous knowledge encoded in the neural network. It is important to note that both the function and the parameters are indispensable. On one hand, if a function has no parameters (e.g., the max pooling function), it can be directly called and there is no need to import it to the knowledge base. On the other hand, as parameters are defined to be bound to a function, parameters themselves are useless if the associated function is missing.

\begin{figure}[!t]
    \centering
    \includegraphics[width=0.85\textwidth]{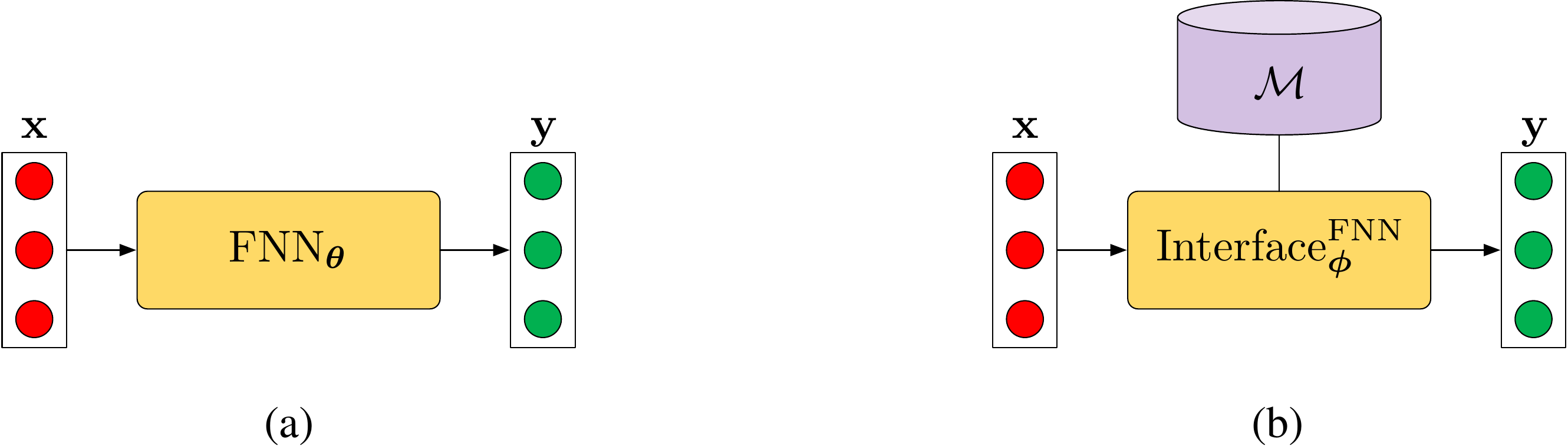}
    \caption{(a) A feed-forward neural network and (b) its interface to the knowledge base. The neural network is represented as a parameterized function $\mathrm{FNN}_{\bm{\theta}}(\cdot)$ that takes $\mathbf{x}$ as input and outputs $\mathbf{y}$. Its interface to the knowledge base $\mathcal{M}$ is also defined as a parameterized function $\mathrm{Interface}^{\mathrm{FNN}}_{\bm{\phi}}(\cdot)$ that shares the same dimensions of input and output with the neural network. An interface is used to facilitate transferring knowledge between a neural network and the knowledge base.} \label{fig:interface}
\end{figure}

\subsection{Continuous Knowledge Base} \label{sec:design_of_a_continuous_knowledge_base}

\subsubsection{Memory Hierarchy}

To build a CKB, we propose to use two levels of real-valued matrices inspired by the use of memory hierarchy in computer architecture \citep{Hennessy:11}. At the high level, the CKB maintains one real-valued matrix $\mathbf{M}^{h}$. At the low level, the CKB maintains $K$ real-valued matrices: $\mathbf{M}^{l}_1, \dots, \mathbf{M}^{l}_k, \dots, \mathbf{M}^{l}_K$. As a result, the CKB consists of $K+1$ real-valued matrices:
\begin{eqnarray}
\mathcal{M} = \left\{ \mathbf{M}^{h}, \mathbf{M}^{l}_1, \dots, \mathbf{M}^{l}_K \right\}
\end{eqnarray}
Note that these real-valued matrices are learnable parameters of the CKB.

\subsubsection{Implementation of an Interface}

To facilitate importing and exporting knowledge between a neural network and the CKB, we introduce an {\em interface} for the neural network. As shown in Figure~\ref{fig:interface}, given a neural network $\mathrm{FNN}_{\bm{\theta}}(\cdot)$, its interface can be defined as a function:
\begin{eqnarray}
\mathbf{y} = \mathrm{Interface}^{\mathrm{FNN}}_{\bm{\phi}}(\mathbf{x}, \mathcal{M}) \label{eq:interface_fnn}
\end{eqnarray}
where $\mathbf{x}$ is the input, $\mathbf{y}$ denotes the output, and $\mathrm{Interface}^{\mathrm{FNN}}_{\bm{\phi}}(\cdot)$ is an interface function parameterized by $\bm{\phi}$  tailored for $\mathrm{FNN}_{\bm{\theta}}(\cdot)$. Note that the input and output of the interface have the same dimensions with those of $\mathrm{FNN}_{\bm{\theta}}(\cdot)$. With interfaces, we can cast importing and exporting knowledge between the neural network $\mathrm{FNN}_{\bm{\theta}}(\cdot)$ and the knowledge base as a {\em function simulation} problem: $\mathrm{Interface}^{\mathrm{FNN}}_{\bm{\phi}}(\mathbf{x}, \mathcal{M})$ runs the same way $\mathrm{FNN}_{\bm{\theta}}(\mathbf{x})$ does given the same input $\mathbf{x}$ (see Sections~\ref{sec:import_knowledge_from_a_neural_network},~\ref{sec:import_knowledge_from_multiple_neural_network}, and~\ref{sec:export_knowledge_to_a_neural_network} for details).

\begin{figure}[!t]
    \centering
    \includegraphics[width=0.55\textwidth]{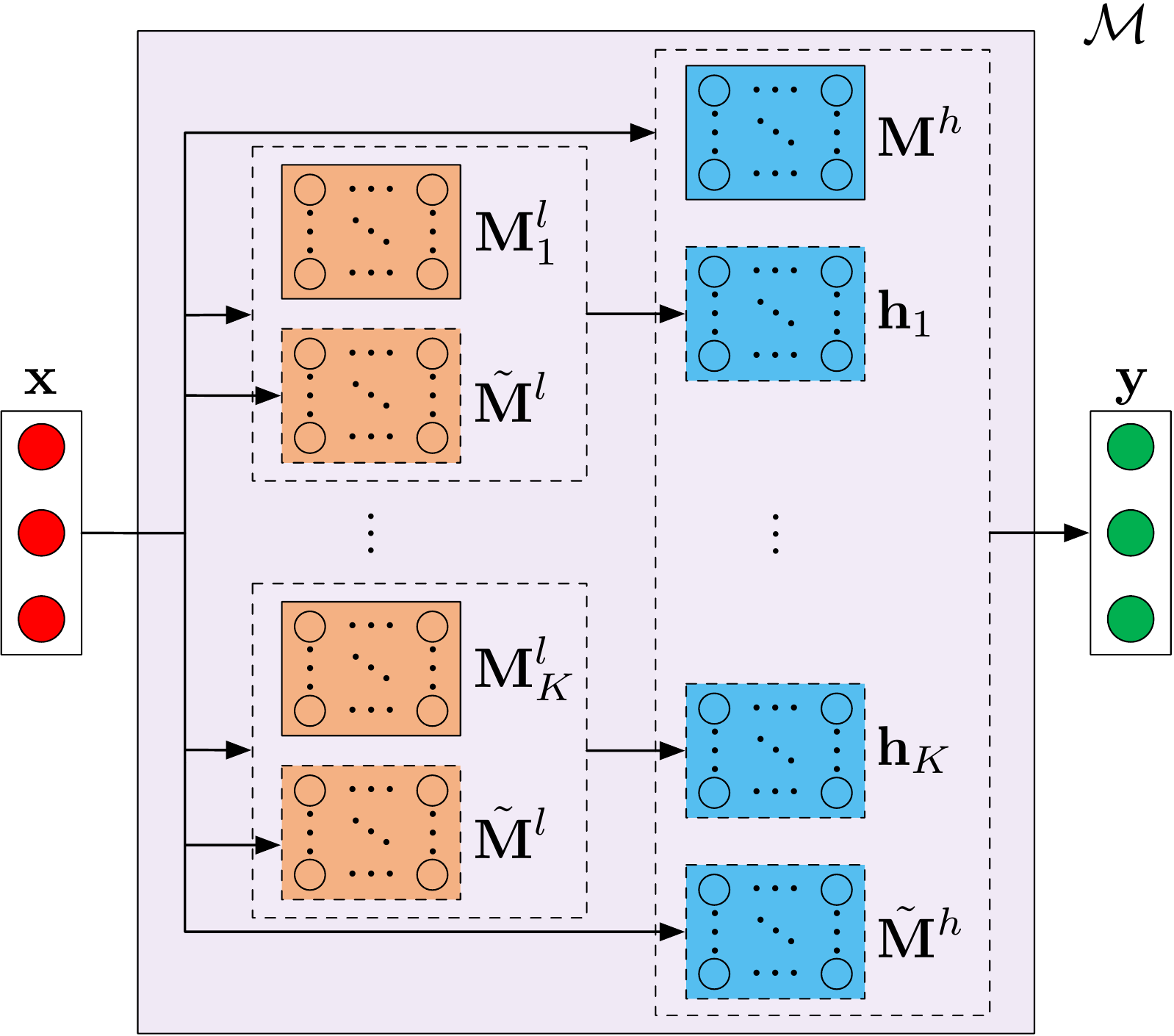}
    \caption{Illustration of how an interface to the continuous knowledge base works. $\mathcal{M}$ is a continuous knowledge base, which is organized as a memory hierarchy: low-level real-valued matrices $\mathbf{M}^{l}_1, \dots \mathbf{M}^{l}_K$ and a high-level matrix $\mathbf{M}^{h}$. The knowledge base provides an interface for each neural network. Given an input $\mathbf{x}$, the interface first generates two extra matrices $\tilde{\mathbf{M}}^{l}$, $\tilde{\mathbf{M}}^{h}$. Note that the matrices generated on the fly are denoted by dashed rectangles. Then, the interface uses the attention function to generate hidden states $\mathbf{h}_1, \dots, \mathbf{h}_{K}$, which are concatenated with $\mathbf{M}^{h}$ and $\tilde{\mathbf{M}}^{h}$ to serve as the key and value (i.e., $\mathbf{H}$) of another attention function to generate the output $\mathbf{y}$.} \label{fig:ckb}
\end{figure}

Figure~\ref{fig:ckb} shows an example that illustrates how an interface works. Given an input $\mathbf{x}$, our approach first adds two extra matrices on the fly:
\begin{align}
\tilde{\mathbf{M}}^{l} &= \mathbf{x} \mathbf{W}^{l}, \\
\tilde{\mathbf{M}}^{h} &= \mathbf{x} \mathbf{W}^{h}
\end{align}
where $\mathbf{W}^{l},\mathbf{W}^{h} \in \bm{\phi}$ are two interface parameters.

Then, for each low-level matrix $\mathbf{M}^{l}_{k}$, we use the attention function \citep{Vaswani:17} to obtain a hidden state:
\begin{equation}
\begin{split}
\tilde{\mathbf{M}}^{l}_{k} &= \big[\mathbf{M}^{l}_{k}; \tilde{\mathbf{M}}^{l}\big], k = 1, \dots, K \\
\mathbf{h}_k &= \mathrm{Attention}(\mathbf{x} \mathbf{W}^{\text{q}}, \tilde{\mathbf{M}}^{l}_k \mathbf{W}^{\text{k}}, \tilde{\mathbf{M}}^{l}_k \mathbf{W}^{\text{v}})
\end{split}
\end{equation}
where $\mathbf{W}^{\text{q}}$, $\mathbf{W}^{\text{k}}$, and $\mathbf{W}^{\text{v}}$ are the transformation matrices for query, key, and value used in the attention mechanism, respectively.

Finally, the output is calculated by calling the attention function again:
\begin{eqnarray}
\mathbf{y} = \mathrm{Attention}(\mathbf{x} \mathbf{W}^{\text{q}}, \mathbf{H} \mathbf{W}^{\text{k}}, \mathbf{H} \mathbf{W}^{\text{v}})
\end{eqnarray}
where the hidden state matrix $\mathbf{H}$ is the concatenation of  the high-level matrix, the hidden states, and the extra matrix $\tilde{\mathbf{M}}^{h}$:
\begin{eqnarray}
\mathbf{H} = \big[\mathbf{M}^{h}; \mathbf{h}_1; \dots; \mathbf{h}_K; \tilde{\mathbf{M}}^{h} \big]
\end{eqnarray}

While every neural network has its own interface to the knowledge base and the interface parameters are often different, the continuous knowledge base $\mathcal{M}$ is shared among all neural networks.

\subsubsection{Interfaces for Global and Local Functions}

As the implementation of an interface is transparent to the model structure, it is easy to define an interface for an arbitrary neural network since one only needs to specify the parameterized function and the dimensions of its input and output.

Often, it is convenient to directly define an interface for the global function if there are only a small number of parameters. For example, the global function for the feed-forward neural network shown in Figure~\ref{fig:fnn} is $\mathrm{FNN}_{\bm{\theta}}(\cdot)$, we can use Eq.~(\ref{eq:interface_fnn}) to define its interface.

However, it is more efficient to define an interface for a local function if the neural network calls it frequently. Consider a recurrent neural network defined as a global function:
\begin{eqnarray}
\mathbf{y} = \mathrm{RNN}_{\bm{\theta}}(\mathbf{x}_{1:T})
\end{eqnarray}
where $\mathbf{x}_{1:T} = \mathbf{x}_1,\dots, \mathbf{x}_T$ is the input sequence and $\mathbf{y}$ is the output.

The global function $\mathrm{RNN}_{\bm{\theta}}(\cdot)$ runs by calling the local function $f_{\bm{\theta}}(\cdot)$ repeatedly:
\begin{eqnarray}
\mathbf{h}_{t} = f_{\bm{\theta}}(\mathbf{x}_t, \mathbf{h}_{t-1}), \ t=1, \dots, T
\end{eqnarray}
where $\mathbf{x}_t$ and $\mathbf{h}_t$ are the input and the hidden state at time step $t$, respectively. Note that we let $\mathbf{y} = \mathbf{h}_T$ for simplicity.

As a result, instead of defining an interface for the global function $\mathrm{RNN}_{\bm{\theta}}(\cdot)$, it is more suitable to define an interface for the local function $f_{\bm{\theta}}(\cdot)$:
\begin{eqnarray}
\mathbf{h}_t = \mathrm{Interface}^{f}_{\bm{\phi}}(\mathbf{x}_t, \mathbf{h}_{t-1}, \mathcal{M})
\end{eqnarray}

\begin{figure}[!t]
    \centering
    \includegraphics[width=0.9\textwidth]{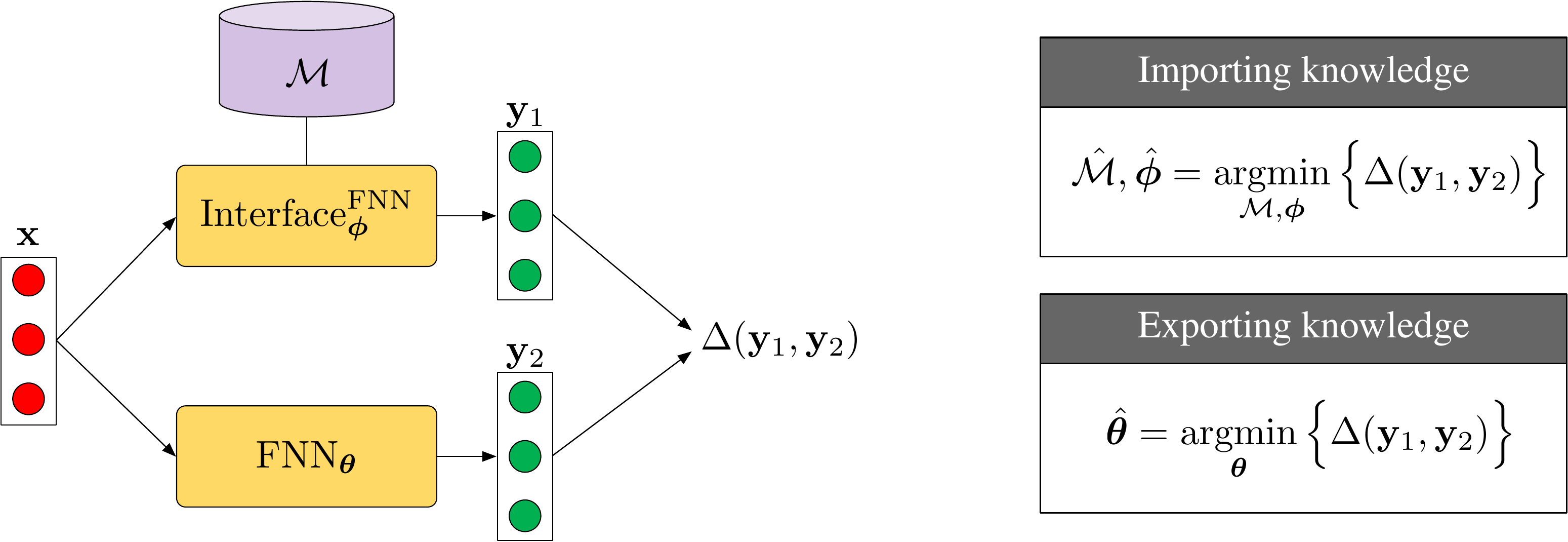}
    \caption{Importing and exporting knowledge for single neural networks. 
    We cast importing knowledge from a neural network to the knowledge base as a function simulation problem: the knowledge is successfully imported only if the interface $\mathrm{Interface}^{\mathrm{FNN}}_{\bm{\phi}}$ runs the same way the neural network $\mathrm{FNN}_{\bm{\theta}}$ does. This is done by finding knowledge base and interface parameters (i.e., $\hat{\mathcal{M}}$ and $\hat{\bm{\phi}}$) that minimize the difference between the outputs of two functions (i.e., $\Delta(\mathbf{y}_1, \mathbf{y}_2)$). Note that the parameters of the neural networks $\bm{\theta}$ are fixed during importing. Similarly, exporting knowledge is also treated as a function simulation problem: finding model parameters $\hat{\bm{\theta}}$ to enable the neural network to imitate the interface while keeping $\mathcal{M}$ and $\bm{\phi}$ fixed.
    } \label{fig:simulation}
\end{figure}

\subsection{Importing Knowledge from Single Neural Networks} \label{sec:import_knowledge_from_a_neural_network}

As shown in Figure~\ref{fig:simulation}, we cast importing knowledge from a neural network to the knowledge base as a {\em function simulation}~\citep{jiao:emnlp20} problem: the knowledge base stores the knowledge encoded in a neural network only if the corresponding interface runs in the same way the neural network does. For example, to import the knowledge from the feed-forward neural network shown in Figure~\ref{fig:fnn} to the knowledge base, we require that the following equation holds for an arbitrary input:
\begin{eqnarray}
\forall \mathbf{x} \in \mathcal{X}: \mathrm{Interface}^{\mathrm{FNN}}_{\bm{\phi}}(\mathbf{x}, \mathcal{M}) = \mathrm{FNN}_{\bm{\theta}}(\mathbf{x})
\end{eqnarray}
where $\mathcal{X}$ is a set of all possible inputs.

As a result, the importing process is equivalent to an optimization problem.
For example, given a set of inputs $\mathcal{D}=\{ \mathbf{x}^{(n)} \}_{n=1}^{N}$, importing $\mathrm{FNN}_{\bm{\theta}}(\cdot)$ to the knowledge base $\mathcal{M}$ is done by
\begin{eqnarray}
\hat{\mathcal{M}}, \hat{\bm{\phi}} = \argmin_{\mathcal{M}, \bm{\phi}} \Big\{ L_{\mathrm{import}}(\mathcal{D}, \mathcal{M}, \bm{\phi}) \Big\}
\end{eqnarray}
where the loss function is defined as
\begin{eqnarray}
L_{\mathrm{import}}(\mathcal{D}, \mathcal{M}, \bm{\phi}) = \sum_{n=1}^{N} \Delta \Big(\mathrm{Interface}^{\mathrm{FNN}}_{\bm{\phi}}(\mathbf{x}^{(n)}, \mathcal{M}), \mathrm{FNN}_{\bm{\theta}}(\mathbf{x}^{(n)}) \Big)
\end{eqnarray}
We use $\Delta(\cdot)$ (e.g., a cosine function) to measure the difference between the outputs of two functions.

\subsection{Importing Knowledge from Multiple Neural Networks} \label{sec:import_knowledge_from_multiple_neural_network}

To import the knowledge encoded in multiple neural networks to the knowledge base, a natural way is to minimize the importing loss functions of these neural networks jointly. For example, let $\mathcal{D}_1 = \{ \mathbf{x}^{(m)} \}_{m=1}^{M}$ be a set of inputs for a feed-forward neural network and $\mathcal{D}_2 = \{\mathbf{x}^{(n)}\}_{n=1}^{N}$ be a set of inputs for a convolutional neural network. Note that the two datasets are independent: the input to the feed-forward neural network can be an image and the input to the convolutional neural network can be a natural language sentence.

The importing loss function of the feed-forward neural network can be defined as
\begin{eqnarray}
L_{\mathrm{import}}^{\mathrm{FNN}}(\mathcal{D}_1, \mathcal{M}, \bm{\phi}_1) = \sum_{m=1}^{M} \Delta \Big( \mathrm{Interface}^{\mathrm{FNN}}_{\bm{\phi}_1} (\mathbf{x}^{(m)}, \mathcal{M}), \mathrm{FNN}_{\bm{\theta}_1}(\mathbf{x}^{(m)}) \Big)
\end{eqnarray}

Similarly, the importing loss function of the convolutional neural network can be defined as
\begin{eqnarray}
L_{\mathrm{import}}^{\mathrm{CNN}}(\mathcal{D}_2, \mathcal{M}, \bm{\phi}_2) = \sum\limits_{n=1}^{N} \Delta \Big( \mathrm{Interface}^{\mathrm{CNN}}_{\bm{\phi}_2} (\mathbf{x}^{(n)}, \mathcal{M}), \mathrm{CNN}_{\bm{\theta}_2}(\mathbf{x}^{(n)})  \Big)
\end{eqnarray}

Then, importing $\mathrm{FNN}_{\bm{\theta}_1}(\cdot)$ and  $\mathrm{CNN}_{\bm{\theta}_2}(\cdot)$ to the knowledge base $\mathcal{M}$ synchronously is given by
\begin{eqnarray}
\hat{\mathcal{M}}, \hat{\bm{\phi}}_1, \hat{\bm{\phi}}_2 = \argmin_{\mathcal{M}, \bm{\phi}_1, \bm{\phi}_2} \Big\{ L_{\mathrm{import}}^{\mathrm{FNN}}(\mathcal{D}_1, \mathcal{M}, \bm{\phi}_1) + L_{\mathrm{import}}^{\mathrm{CNN}}(\mathcal{D}_2, \mathcal{M}, \bm{\phi}_2) \Big\}
\end{eqnarray}
It is easy to extend the above approach to more than two neural networks.

\subsection{Exporting Knowledge to a Neural Network} \label{sec:export_knowledge_to_a_neural_network}

We can also use the interface to export the knowledge stored in the CKB to a neural network. Still, we treat exporting knowledge from the knowledge base to a neural network as a function simulation problem: the knowledge is exported to the neural network only if the the neural network runs in the same way the interface does. For example, to export the knowledge from the CKB to the feed-forward neural network shown in Figure~\ref{fig:fnn}, we require that the following equation holds for an arbitrary input:
\begin{eqnarray}
\forall \mathbf{x} \in \mathcal{X}: \mathrm{FNN}_{\bm{\theta}}(\mathbf{x}) = \mathrm{Interface}^{\mathrm{FNN}}_{\bm{\phi}}(\mathbf{x}, \mathcal{M})
\end{eqnarray}

As a result, the importing processing is also equivalent to an optimization problem: our goal is to modify the parameters of the neural network to minimize the difference between the neural network and its interface. For example, given a set of inputs $\mathcal{D}=\{ \mathbf{x}^{(n)} \}_{n=1}^{N}$, exporting the knowledge stored in the knowledge base $\mathcal{M}$ to $\mathrm{FNN}_{\bm{\theta}}(\cdot)$ is done by
\begin{eqnarray}
\hat{\bm{\theta}} = \argmin_{\bm{\theta}}\Big\{ L_{\mathrm{export}}(\mathcal{D}, \bm{\theta}) \Big\}
\end{eqnarray}
where the loss function is defined as
\begin{eqnarray}
L_{\mathrm{export}}(\mathcal{D}, \bm{\theta}) = \sum_{n=1}^{N} \Delta \left(\mathrm{Interface}^{\mathrm{FNN}}_{\bm{\phi}}(\mathbf{x}^{(n)}, \mathcal{M}), \mathrm{FNN}_{\bm{\theta}}(\mathbf{x}^{(n)}) \right)
\end{eqnarray}

As the knowledge base and interfaces are fixed during the exporting process, exporting knowledge to multiple neural networks is equivalent to exporting it to single neural networks in parallel.

\section{Experiments}

\subsection{Setting}
We evaluated our approach on text classification on two public datasets: 
\begin{enumerate}
\setlength{\itemsep}{4pt}
\setlength{\parskip}{0pt}
\setlength{\parsep}{0pt}
\item Amazon positive-negative review dataset \citep{Fu:18}. The training set contains 796,000 reviews. The test set contains 4,000 reviews. We split the original test set into two parts: 2,000 reviews as the validation set and 2,000 reviews as the test set. On average, each review contains 19 words.
\item Yelp polarity review dataset \citep{Zhang:15}. The training set contains 560,000 reviews. The test set contains 38,000 reviews. We split the original training set into two parts: 550,000 reviews as the training set and 10,000 as the validation set. On average, each review contains 163 words.
\end{enumerate}

\begin{table}[htb]
\centering
\scalebox{1.0}{
\begin{tabular}{l|c|c|c||c|c}
\Bhline
\multirow{2}{*}{\textbf{Configuration}}  & \multicolumn{3}{c||}{\textbf{\#Param.}} & \multicolumn{2}{c}{\textbf{Accuracy}} \\
\cline{2-6}
    &  model ($\bm{\theta}$) & interface ($\bm{\phi}$) & CKB ($\mathcal{M}$) & Amazon & Yelp \\
    \hline \hline
    RNN  & ~~0.20M & - & - & 84.30 & 96.31\\
    RNN $\mapsto$ CKB $\mapsto$ RNN &  - & ~~1.05M & 81.92K & 84.40 & 96.32 \\
    \hline
    CNN  & ~~0.26M & - & - & 84.35 & 95.95 \\
    CNN $\mapsto$ CKB $\mapsto$ CNN & - & ~~0.79M & 81.92K & 84.15 & 95.67 \\
    \hline
ANN  & ~~0.79M & - & - & 84.15 & 93.51 \\
ANN $\mapsto$ CKB $\mapsto$ ANN & - & ~~1.31M & 81.92K & 84.15 & 93.61 \\
    \hline
    BERT & 85.64M & - & - & 87.90 & 97.34 \\
    BERT $\mapsto$ CKB $\mapsto$ BERT & - & 19.09M & ~~1.70M & 87.60 & 96.69 \\
    \hline
    GPT-2 & 85.64M & - & - & 87.80 & 97.62 \\
    GPT-2 $\mapsto$ CKB $\mapsto$ GPT-2 & - & 19.09M & ~~1.70M & 87.75 & 97.13 \\
    \Bhline
\end{tabular}
}
\caption{Results of importing and exporting knowledge for single models. ``RNN $\mapsto$ CKB $\mapsto$ RNN'' denotes first importing the knowledge encoded in the RNN model that achieves an accuracy of 84.30 on the Amazon dataset to CKB, from which the knowledge stored is exported to another RNN model with the same model structure. Note that the number of parameters does not include the word embedding layer and the classification layer.} \label{tab:single_model_exp}
\end{table}

\def \rnn{RNN}
\def \cnn{CNN}
\def \ann{ANN}

We used the following five neural networks tailored for text classification in our experiments:
\begin{enumerate}
\setlength{\itemsep}{4pt}
\setlength{\parskip}{0pt}
\setlength{\parsep}{0pt}
    \item \rnn{}~\citep{textrnn:ijcai16}: recurrent neural network. We used a single gate recurrent unit (GRU)~\citep{gru:emnlp14} layer as the encoder. Its hidden size is set to 256. We defined the interface for RNN at the local level: the input of the interface consists of the $t$-th word $\mathbf{x}_t$ and the $(t-1)$-th hidden state $\mathbf{h}_{t-1}$ and the output is the $t$-th hidden state $\mathbf{h}_t$. 
    \item \cnn{}~\citep{textcnn:emnlp14}: convolution neural network. We followed~\cite{textcnn:emnlp14} to use three filter windows (i.e., 3, 4, and 5) with 80 feature maps, respectively. We defined its interface at the local level: the input of the interface is a sequence of consecutive words $\mathbf{X}_{t:t+w}$ and the output is a hidden state $\mathbf{h}_t$.
    \item \ann{}~\citep{Vaswani:17}: attention-based neural network. We used a single Transformer encoder layer as the encoder. Its hidden size and intermediate size are 256 and 1024, respectively. We defined the interface for ANN at the global level: the input of the interface is the entire sequence $\mathbf{X}$ and the output is the sequence of hidden states $\mathbf{H}$. 
    \item BERT~\citep{Devlin:19}: Bidirectional encoder representations from Transformers. The BERT base model~\footnote{\url{https://huggingface.co/bert-base-uncased}} was fine-tuned on the two text classification datasets. We defined the interface for BERT at the global level: the input of the interface is the entire sequence $\mathbf{X}$ and the output is the final output of the BERT encoder. Note that the BERT base model uses 12 attention layers. Our interface directly predicts the output of the 12-th layer.
    \item GPT-2~\citep{gpt2:19}: a language model based on masked self-attention. We fine-tuned GPT-2~\footnote{\url{https://huggingface.co/gpt2}} for text classification and defined its interface similar to that of BERT.
    \item ALBERT~\citep{albert:iclr20}: a lite BERT.~\footnote{\url{https://huggingface.co/albert-base-v2}.} The interface for ALBERT is defined at the global level in a similar way.
\end{enumerate}

We removed documents that contain less than 5 words and only retained the first 512 tokens for documents that have more than 512 tokens. For \rnn{}, \cnn{}, and \ann{}, we used BPE~\citep{bpe:acl16} with 32K operations to preprocess the datasets. For other methods, we used their built-in tokenizers to preprocess the datasets.

We used 10 low-level $\text{30}\times\text{256}$ matrices and one high-level $\text{20}\times\text{256}$ matrix to build the CKB for small models (i.e., \rnn{}, \cnn{}, and \ann{}). For big models like BERT and GPT-2, the CKB contains 20 low-level $\text{40}\times\text{2,048}$ metrices and one high-level $\text{20} \times \text{2,048}$ matrix. We used AdamW~\citep{adamw:iclr19} to optimize parameters of the CKB, interfaces, and neural networks. Please refer to Appendix~\ref{exp_config} for more details.

\begin{table}[t!]
\centering
\scalebox{1.0}{
\begin{tabular}{l|l||c}
    \Bhline
    \textbf{Method}  & \textbf{Configuration} & \multicolumn{1}{c}{\textbf{Accuracy}} \\
    \hline \hline
    \multirow{3}{*}{\textit{Single}} & \rnn{} & 84.30 \\
    \cline{2-3}
    & \cnn{} & 84.35 \\
    \cline{2-3}
    & \ann{} & 84.15 \\
    \hline
    \multirow{6}{*}{\textit{Ours}} & \rnn{} $\mapsto$ CKB $\mapsto$ \cnn{} & 83.35 \\
    \cline{2-3}
    & \rnn{} $\mapsto$ CKB $\mapsto$ \ann{} & 83.10 \\
    \cline{2-3}
    & \cnn{} $\mapsto$ CKB $\mapsto$ \rnn{} & 82.70 \\
    \cline{2-3}
    & \cnn{} $\mapsto$ CKB $\mapsto$ \ann{} & 83.85 \\
    \cline{2-3}
    & \ann{} $\mapsto$ CKB $\mapsto$ \rnn{} & 82.05 \\
    \cline{2-3}
    & \ann{} $\mapsto$ CKB $\mapsto$ \cnn{} & 83.10
    \\
    \Bhline
\end{tabular}
}
\caption{Results of knowledge transferring between different models on the Amazon dataset.} \label{tab:diff_model_exp}
\end{table}

\begin{table}[t!]
\begin{center}
\scalebox{1.0}{
    \begin{tabular}{l|l||c}
    \Bhline
    \textbf{Method} & \textbf{Configuration} & \textbf{Accuracy} \\
    \hline \hline
    \multirow{5}{*}{\textit{Single}} & \rnn{}  & 84.30 \\
    \cline{2-3}
    & \cnn{} &  84.35 \\
    \cline{2-3}
    & \ann{} &  84.15 \\
    \cline{2-3}
    & BERT &  87.90 \\
    \cline{2-3}
    & GPT-2 &  87.80 \\
    \hline
    \multirow{5}{*}{\textit{Ensemble}} & \rnn{} \& \cnn{} & 85.25 \\
    \cline{2-3}
    & \cnn{} \& \ann{} &  85.40 \\
    \cline{2-3}
    & \ann{} \& \rnn{} &  85.00 \\
    \cline{2-3}
    & \rnn{} \& \cnn{} \& \ann{} & 85.35 \\
    \cline{2-3}
    & BERT \& GPT-2 &  88.45 \\
    \hline
    \multirow{11}{*}{\textit{Ours}} & $\{$\rnn{}, \cnn{}$\} \mapsto$ CKB $\mapsto$ \rnn{}  & 85.05 \\
    \cline{2-3}
    & $\{$\rnn{}, \cnn{}$\} \mapsto$ CKB $\mapsto$ \cnn{}  & 84.05 \\
    \cline{2-3}
    & $\{$\cnn{}, \ann{}$\} \mapsto$ CKB $\mapsto$ \cnn{}  & 84.25 \\
    \cline{2-3}
    & $\{$\cnn{}, \ann{}$\} \mapsto$ CKB $\mapsto$ \ann{} & 84.30 \\
    \cline{2-3}
    & $\{$\ann{}, \rnn{}$\} \mapsto$ CKB $\mapsto$ \ann{}  & 84.35 \\
    \cline{2-3}
    & $\{$\ann{}, \rnn{}$\} \mapsto$ CKB $\mapsto$ \rnn{}  & 85.20 \\
    \cline{2-3}
    & $\{$\rnn{}, \cnn{}, \ann{}$\} \mapsto$ CKB $\mapsto$ \rnn{} &  84.95 \\
    \cline{2-3}
    & $\{$\rnn{}, \cnn{}, \ann{}$\} \mapsto$ CKB $\mapsto$ \cnn{} &  83.95 \\
    \cline{2-3}
    & $\{$\rnn{}, \cnn{}, \ann{}$\} \mapsto$ CKB $\mapsto$ \ann{} &  84.10 \\
    \cline{2-3}
    & $\{$BERT, GPT-2$\} \mapsto$ CKB $\mapsto$ BERT &  88.20 \\
    \cline{2-3}
    & $\{$BERT, GPT-2$\} \mapsto$ CKB $\mapsto$ GPT-2 &  88.35 \\
    \Bhline
    \end{tabular}
}
\end{center}
\caption{Results of importing and exporting knowledge for multiple models on the Amazon dataset.} \label{tab:multi_model_exp}
\end{table}

\subsection{Importing and Exporting Knowledge for Single Neural Networks}

Table~\ref{tab:single_model_exp} shows the results of importing and exporting knowledge for single neural networks. This experiment aims to verify whether CKB is able to import and export continuous knowledge. We find that our approach is capable of retaining the expressive power of the original neural network across a variety of architectures.

Table~\ref{tab:single_model_exp} also lists the numbers of model, interface, and CKB parameters. The interface for CNN has the fewest parameters (i.e., 0.79M) because it only takes a substring of the input sequence as input. As the interface for \rnn{} takes both the current token and the last hidden state as input, it has more parameters (i.e., 1.05M) than that of CNN. The interface for \ann{} has more parameters than that of \rnn{} because it takes the entire sequence as input and contains an additional feed-forward layer. Since the interfaces for BERT and GPT-2 directly imitate the ouput of the final layer (i.e., the 12-th layer), they have much fewer parameters than the original models.

Table~\ref{tab:diff_model_exp} shows the results of importing and exporting knowledge for different single models. We import the knowledge from one model to the CKB and then export the knowledge to a different model, which is similar to the setting of zero-shot learning. We find that the knowledge stored in \rnn{}, \cnn{}, and \ann{} can be transferred to each other with only small performance degradation.

\subsection{Importing and Exporting Knowledge for Multiple Neural Networks}

Table~\ref{tab:multi_model_exp} shows the results of importing and exporting knowledge for multiple models on the Amazon dataset. We find that first importing multiple models to the CKB and then exporting the fused knowledge to a single model can result in improved accuracy for the single model. For example, the single model RNN obtains an accuracy of 84.30\% while ``\{RNN, CNN\} $\mapsto$ CKB $\mapsto$RNN'' achieves 85.05\%, suggesting that the knowledge stored in CKB helps to improve RNN. Similar results were also observed for larger models such as BERT and GPT-2. However, we also found that not all single models to which the knowledge is exported obtain higher accuracies. For example, ``\{RNN, CNN\} $\mapsto$ CKB $\mapsto $CNN'' obtains a lower accuracy than the original CNN. As a result, how to ensure all participating models benefit from the integration still needs further exploration.

Our approach significantly differs from model ensemble in two aspects. First, while model ensemble has to maintain all participating models during inference, CKB can export its knowledge to a single model. Second, model ensemble requires all participating models are trained for the same task while CKB in principle can take advantage of models trained for different tasks.

\subsection{Effect of the Capacity of Continuous Knowledge Base}

\begin{table}[ht]
\centering
\scalebox{1.0}{
    \begin{tabular}{l|c||c}
    \Bhline
        \textbf{Configuration} & \textbf{\#Param.} & \textbf{Accuracy} \\
    \hline \hline
        BERT & - & 87.90 \\
    \hline
        \multirow{3}{*}{BERT $\mapsto$ CKB $\mapsto$ BERT} & 81.92K & 86.55 \\
    \cline{2-3}
        & ~~1.70M & 87.60 \\
    \cline{2-3}
        & ~~3.11M & 87.65 \\
    \Bhline
    \end{tabular}
}
\caption{Effect of the capacity of the continuous knowledge base on the Amazon dataset.} \label{tab:eff_capacity}
\end{table}

Table \ref{tab:eff_capacity} shows the effect of model capacity of CKB on classification accuracy. We find that the accuracy generally rises with the increase of the number of model parameters (i.e., $\mathcal{M}$) and the benefit becomes modest on larger models. 

\subsection{Knowledge Distillation via Continuous Knowledge Base}

\begin{table}[htb]
\begin{center}
\scalebox{1.0}{
   \begin{tabular}{l|c|c||c}
   \Bhline
   \multicolumn{2}{c|}{\multirow{1}{*}{\textbf{Method}}} & \multirow{1}{*}{\textbf{Hidden Size}} & \multicolumn{1}{c}{\textbf{Accuracy}} \\
   \hline \hline
   \multirow{2}{*}{\textit{Base}} & \rnn{} & 256 & 84.30 \\
   \cline{2-4}
   & \rnn{} & ~~64 & 83.30 \\
   \hline
   \textit{KD} & \rnn{} & ~~64 & 83.00 \\
   \hline
   \textit{Ours} & \rnn{} & ~~64 & 83.00 \\
   \Bhline
   \end{tabular}
}
\end{center}
\caption{Results of mimicking knowledge distillation (KD) by CKB on the Amazon dataset.} \label{tab:kd_exp}
\end{table}

We can use the CKB to realize the goal of knowledge distillation (KD). As our CKB can export the stored knowledge to a blank model, KD can be done by importing the knowledge from the teacher model to the CKB and then exporting the knowledge to the student model. In our experiments, we used a big \rnn{} with 256 hidden size and a small \rnn{} with 64 hidden size as the teacher and the student models, respectively. As shown in Table~\ref{tab:kd_exp}, KD with our CKB achieved the same performance as the standard KD method. The performances of two KD methods are slightly worse than that of the small model trained on labeled data. One possible reason is that the teacher model contains noise, affecting the performance of the student model.

\subsection{Transfer Learning via Continuous Knowledge Base}

\begin{table}[!htb]
\begin{center}
\scalebox{1.0}{
   \begin{tabular}{c|l|c||c}
   \Bhline
   \multicolumn{2}{c|}{\multirow{1}{*}{\textbf{Method}}} & \multirow{1}{*}{\textbf{Setting}} & \multicolumn{1}{c}{\textbf{Accuracy}} \\
   \hline \hline
   \multirow{2}{*}{\textit{PT \& FT}} & ALBERT & Init $+$ FT & 79.20 \\
   \cline{2-4}
   & ALBERT & PT $+$ FT & 84.50 \\
   \hline
   \multirow{2}{*}{\textit{Ours}} & CKB & Init $+$ FT & 82.10 \\
   \cline{2-4}
   & CKB & TL $+$ FT & 84.10 \\
   \Bhline
   \end{tabular}
}
\end{center}
\caption{Results of Transfer Learning by CKB on the Amazon dataset. ``PT'', ``FT'', and ``TL'' denote pre-training, fine-tuning, and transfer learning, respectively. ``Init'' means model initialization.} \label{tab:tl_exp}
\end{table}

It is easy to imitate transfer learning based on our CKB. In our experiments, we used the CKB to mimic the transfer learning where a pre-trained language model (i.e., ALBERT) is fine-tuned for text classification. As shown in Table~\ref{tab:tl_exp}, the ALBERT trained from scratch on the Amazon dataset obtained 79.20 accuracy scores on the test set while the pre-trained ALBERT can obtain 84.50 accuracy scores after fine-tuning. Analogously, the CKB-based model trained from scratch with a randomly initialized CKB on the Amazon dataset achieved 82.10 accuracy scores on the test set. However, the CKB-based model, which imported the knowledge from the pre-trained ALBERT, performed much better. Note that in the knowledge transfer phase, we only used the unlabeled text data from the Amazon dataset to import the knowledge from the pre-trained ALBERT to the CKB.

\section{Related Work}
Our work draws inspiration from two lines of research: memory networks and knowledge bases from pre-trained models.

\subsection{Memory Networks}
Memory networks (MNs) are first proposed by~\citet{memory:iclr15}. The proposed MNs reason with inference components combined with a long-term memory which acts as a dynamic knowledge base to store some knowledge from the input. \citet{e2e-memory:nips15} extend MNs to the end-to-end paradigm by introducing the attention mechanism~\citep{rnnsearch:iclr15} to estimate the relevance of each item in the memory. \citet{dmns:icml16} propose dynamic memory networks (DMNs) that use episodic memories to help generate better answers to given questions. The episodic memory in DMNs can be updated dynamically according to the input. \citet{nematzadeh:iclr20} also argue that the separation of computation and storage is necessary and discuss the advantage of improving memory in AI systems. Different from the existing MNs which store the input-related knowledge, the proposed CKB is a global knowledge base that aims to store the knowledge from different competent neural network models.

\subsection{Knowledge Bases from Pre-trained Models}
Recently, a number of works have been studied on what does the pre-trained language model learns \citep{fabio:emnlp19,bouraoui:aaai20,rogers:arxiv20,wang:arxiv20}. \citet{fabio:emnlp19} convert the fact (i.e., subject-relation-object triple) into the cloze statement to test the factual and commonsense knowledge in the pre-trained language model. By transforming relational triples into masked sentences, \citet{Feldman:emnlp19} propose to mine commonsense knowledge from pre-trained models. \citet{bouraoui:aaai20} fine-tune the pre-trained BERT~\citep{Devlin:19} to predict whether a given word pair is likely to be an instance of some relations. \citet{wang:arxiv20} state that pre-trained language models would be open knowledge graphs and propose an unsupervised method to build knowledge graphs. \citet{net2net:neurips20} propose a conditional invertible neural network to translate between fixed representations from different off-the-shelf models. These methods show that neural network models would contain knowledge while our CKB investigates how to store and use this uninterpretable knowledge.

\section{Conclusion and Future Work}
We propose to build a universal continuous knowledge base (CKB) in this work. Different from conventional knowledge bases using discrete symbols to represent information, the proposed CKB stores the knowledge in multi-level real-valued matrices. Based on the formalization where a neural network model is a parameterized composite function that maps the input to the output, our CKB imports the knowledge from the neural network model by learning the mapping between the input and the output with the model-dependent interface. Experiments on text classification show that continuous knowledge can be imported and exported between neural networks and the CKB. 
Our CKB can also mimic knowledge distillation and transfer learning in a novel paradigm. In the future, we will extend the CKB to cross-task scenarios.

Our work has only touched the surface of building a universal continuous knowledge base. There are a number of interesting directions awaiting further exploration: a more sophisticated design of memory hierarchy, integrating neural networks trained for different AI tasks, continual learning that can import multiple neural networks asynchronously, visualization and interpretation of the internal workings, and importing and exporting knowledge between discrete and continuous KBs. 

\section*{Acknowledgements}
We would like to thank Qun Liu, Xin Jiang, and Meng Zhang for their constructive comments on this work. This
work was supported by the National Key R\&D
Program of China (No. 2017YFB0202204), National Natural Science Foundation of China (No. 61925601, No. 61761166008) and Huawei Noah's Ark Lab.

\bibliographystyle{apalike}
\bibliography{refs}

\clearpage

\appendix
\section{Experimental Details} \label{exp_config}

\subsection{Hyper-parameter Values}

When importing and exporting continuous knowledge between neural networks and the knowledge base, each mini-batch contains 8,192 tokens for the Amazon dataset and 24,576 tokens for the Yelp dataset, respectively. We used the AdamW optimizer~\citep{adamw:iclr19} with $\beta_1=0.9$, $\beta_2=0.98$, $\epsilon=10^{-9}$, and L2 weight decay of $0.01$ to optimize parameters. For knowledge importing, we set the learning rate to 2e-4. For knowledge exporting, the learning rate was set to 1e-4.

\subsection{Model Selection}

For importing and exporting knowledge for single neural networks, model selection is done by choosing the checkpoint with the highest accuracy on the validation set. When synchronously importing knowledge from multiple neural networks to the continuous knowledge base (CKB), a problem is that these models might not achieve the highest performance on the validation set at the same time.
To address this problem, we select the checkpoint as follows: 
\begin{eqnarray}
    \hat{c} = \argmax_{c}\bigg\{\! \min_{i \in [1, N ]}\!\Big\{\mathrm{acc}(c, \mathrm{NN}_i) \Big\} \bigg\}
\end{eqnarray}
where $N$ is the number of neural networks, $\mathrm{NN}_i$ is the $i$-th neural network, $c$ is a checkpoint, and $\mathrm{acc(\cdot)}$ is a function that calculates accuracy on the validation set.

\subsection{Runtime Environment}
We conducted all experiments on a server with the following environment:
\begin{itemize}
    \item Operation System: Ubuntu 18.04.2 LTS
    \item CPU: AMD EPYC 7302 16-Core Processor
    \item GPU: GeForce RTX 3090
\end{itemize}
\end{document}